\pdfoutput=1

\documentclass[11pt]{article}

\usepackage{ACL2023}
\usepackage{tablefootnote}
\usepackage{times}
\usepackage{latexsym}
\usepackage{diagbox}
\usepackage[T1]{fontenc}

\usepackage[utf8]{inputenc}

\usepackage{microtype}

\usepackage{inconsolata}

\usepackage{graphicx}

\title{Towards Data-efficient Customer Intent Recognition with Prompt-based Learning Paradigm}

\author{
    \textbf{Hengyu Luo\textsuperscript{1*}},  
    \textbf{Peng Liu\textsuperscript{2}},  
    \textbf{Stefan Esping\textsuperscript{2}} \\
    \textsuperscript{1}University of Helsinki, Finland \\
    \textsuperscript{2}Ingka Group, IKEA, Malmö, Sweden \\
    \small{
        \textbf{Correspondence:} \href{mailto:peng.liu4@ingka.ikea.com}{peng.liu4@ingka.ikea.com}
    } \\
    \small{\textsuperscript{*}Work done while at Uppsala University.}
}

\begin{document}
\maketitle
\begin{abstract}
Recognizing customer intent accurately with language models based on customer-agent conversational data is essential in today's digital customer service marketplace, but it is often hindered by the lack of sufficient labeled data. In this paper, we introduce the prompt-based learning paradigm that significantly reduces the dependency on extensive datasets. Utilizing prompted training combined with answer mapping techniques, this approach allows small language models to achieve competitive intent recognition performance with only a minimal amount of training data. Furthermore, We enhance the performance by integrating active sampling and ensemble learning strategies in the prompted training pipeline. Additionally, preliminary tests in a zero-shot setting demonstrate that, with well-crafted and detailed prompts, small language models show considerable instruction-following potential even without any further training. These results highlight the viability of semantic modeling of conversational data in a more data-efficient manner with minimal data use, paving the way for advancements in AI-driven customer service.
\end{abstract}

\section{Introduction}

In the contemporary era marked by the digitalization of customer services across various industrial sectors, there is heightened attention to the analysis of customers' intents during their interactions with service agents. The sheer volume of conversational data necessitating analysis has underscored the importance of leveraging language models for a quick and automatic recognition of customer intents.

Previously, under the traditional fine-tuning paradigm, it has been common practice to utilize labeled datasets to fine-tune existing pretrained models, thereby enhancing their adaptability to downstream tasks. For instance, in the domain of customer intent recognition, conversational data labeled with various customer intents are used to fine-tune pretrained models to better align with intent recognition tasks. However, this approach encounters two significant challenges in the industrial context. Firstly, the process of obtaining labeled data is inherently labor-intensive, necessitating a significant investment in human labor and associated costs to secure high-quality annotations. Secondly, because of the operational overhead associated with large language models (LLMs), there is a preference within the industry for small language models (SLMs) that offer greater flexibility and are easier to deploy. Nevertheless, traditional fine-tuning often anticipates larger language models and more extensive annotated data to achieve satisfactory performance in downstream tasks, posing a challenge to meet these industrial requirements.

In this context, the prompt-based learning paradigm, particularly the concept of prompted training (also known as instruction tuning), has recently emerged as a viable alternative to traditional fine-tuning \citep{liu2023pre}. This approach demonstrates the potential to achieve competitive performance with a reduced requirement for labeled samples. 

In this paper, we investigate the efficacy of the prompt-based learning paradigm in recognizing customer intents, drawing on customer-agent conversational data in the retail customer service domain, especially when paired with SLMs. In light of the challenges highlighted earlier, the core contributions and findings of this paper are:

\begin{itemize}
\item Utilizing prompted training combined with answer mapping techniques, we demonstrate the potential of SLMs in achieving notable intent recognition performance in retail interactions with minimal training data involved. This represents a significant advancement over traditional fine-tuning methods in terms of data efficiency.
\item We propose the integration of active sampling and ensemble learning strategies within the prompt-based learning framework, which further contributes to an enhancement in intent recognition performance in minimal-data scenarios.
\item In zero-shot settings, by leveraging carefully crafted and detailed prompts, SLMs like FLAN-T5-large still show considerable instruction-following potential. Without any training data, these models demonstrate promising intent recognition performance.
\end{itemize}

With these contributions in perspective, the remainder of this paper is structured as follows: We introduce related work in Section 2. Section 3 offers a brief introduction to the dataset we utilize. Section 4 elaborates our prompt-based learning pipeline, including methodologies and experimental setups. Section 5 presents our findings and an in-depth analysis. We conclude with insights and potential directions for future research in Section 6. Through our experiments, we aim to uncover a more data-efficient strategy for deploying prompt-based learning with SLMs, thus enhancing the semantic modeling of conversational data and gaining advancements in AI-driven customer service with minimal data use.

\section{Related Work}

The recent advent of pre-trained language models, such as BERT \citep{devlin2018bert}, GPT \citep{radford2018gpt}, and RoBERTa \citep{liu2019roberta}, has brought about transformative changes in NLP. These models, pre-trained on vast text repositories, have captured intricate language patterns and representations, enabling them to be fine-tuned for a variety of downstream tasks. However, traditional fine-tuning approaches require a significant amount of labeled data, which can be a constraint, especially in domain-specific applications.

In the realm of prompt-based learning paradigm, recent models such as PET-TC \citep{schick-schutze-2021-exploiting}, PET-Gen \citep{schick2020few}, and LM-BFF \citep{gao-etal-2021-making} have demonstrated the potential of prompt-based learning. For instance, LM-BFF has introduced innovative techniques like automated template generation and label word selection, significantly reducing the need for manual prompt engineering \citep{gao-etal-2021-making}. These models hold promise for domain-specific text classification, a focus of our research, where acquiring extensive labeled data can be challenging. Conversely, in zero-shot settings, models like LAMA \citep{petroni-etal-2019-language} and GPT-3 \citep{brown2020language} have explored tuning-free prompting. This approach relies on the careful engineering of prompts for achieving desired outputs.

\section{Dataset}

The dataset, sourced from a furniture retail company's customer support service, captures conversations between customers and agents over 3 months. All data within the set is text-based and has been cleaned to exclude any personally identifiable information (PII), safeguarding the privacy of both customers and agents. This collection consists of 7,477  conversations between a consumer and an agent.

Following the data acquisition, the data were manually labeled with 13 distinct labels reflecting various customer intents, plus an "other" label for instances not aligning with the predefined 13 labels. These intents cover topics such as order processing, product inquiry, billing issues, and technical support. A demonstration sample is shown in Table \ref{data-example}. The dataset exhibits an imbalance: certain intentions are notably more frequent than others. The distribution of labels can be seen in table \ref{label}.

\begin{table*}
\centering
{%
\begin{tabular}{ll}
\hline
\textbf{Conversation} &
  \textbf{Label} \\ \hline
\begin{tabular}[c]{@{}l@{}}...\\ Customer: I'd like to know the detailed instruction to assemble this desk. \\ Agent: Sure, let me check our website. Hold on, please.\\ ...\end{tabular} &
  “Help integrating the product” \\ \hline
\end{tabular}%
}
\caption{An example of the labeled customer-agent conversation in the dataset}
\label{data-example}
\end{table*}

\begin{table}
\centering
\begin{tabular}{lll}
\hline
\textbf{No.} & \multicolumn{1}{l}{\textbf{Name of Label}}   & \multicolumn{1}{l}{\textbf{Count}} \\ \hline
0   & Product / Service Availability        & 951                                  \\
1   & General                             & 113                                  \\
2   & General after Purchase              & 108                                  \\
3   & Help Integrating the Product  & 306                                  \\
4   & Initiate After-sales Service         & 907                                  \\
5   & Issue Handling                & 655                                  \\
6   & Order Creation                      & 1259                                 \\
7   & Order Fulfillment Issues            & 997                                  \\
8   & Order Processing                    & 1069                                 \\
9  & Other                               & 102                                  \\
10  & Planning \& Advice                  & 192                                  \\
11  & Prepare for Exchange \& Returns     & 531                                  \\
12  & Product / Service Information       & 283                                  \\
13  & Service Fulfillment  & 29                                   \\ \hline
total & & 7477\\ \hline
\end{tabular}
\caption{Label distribution in the dataset}
\label{label}
\end{table}

The dataset is split into three subsets with stratified sampling for training use: a combined training and development set (50\% of the total, 3738 entries), a validation set (25\%, 1870 entries), and a test set (25\%, 1869 entries). The test set only serves for the final evaluation, to ensure that no insights from the test set are revealed from prior experiments and offer an objective evaluation of the performance. All the experimental results are based on the test set.

\section{Methodology and Experimental Setup}

\subsection{Prompt-based Learning Pipeline: Prompting, Predicting, Answer Mapping}

\begin{figure*}[hbtp]
    \centering
    \includegraphics[width=0.9\textwidth]{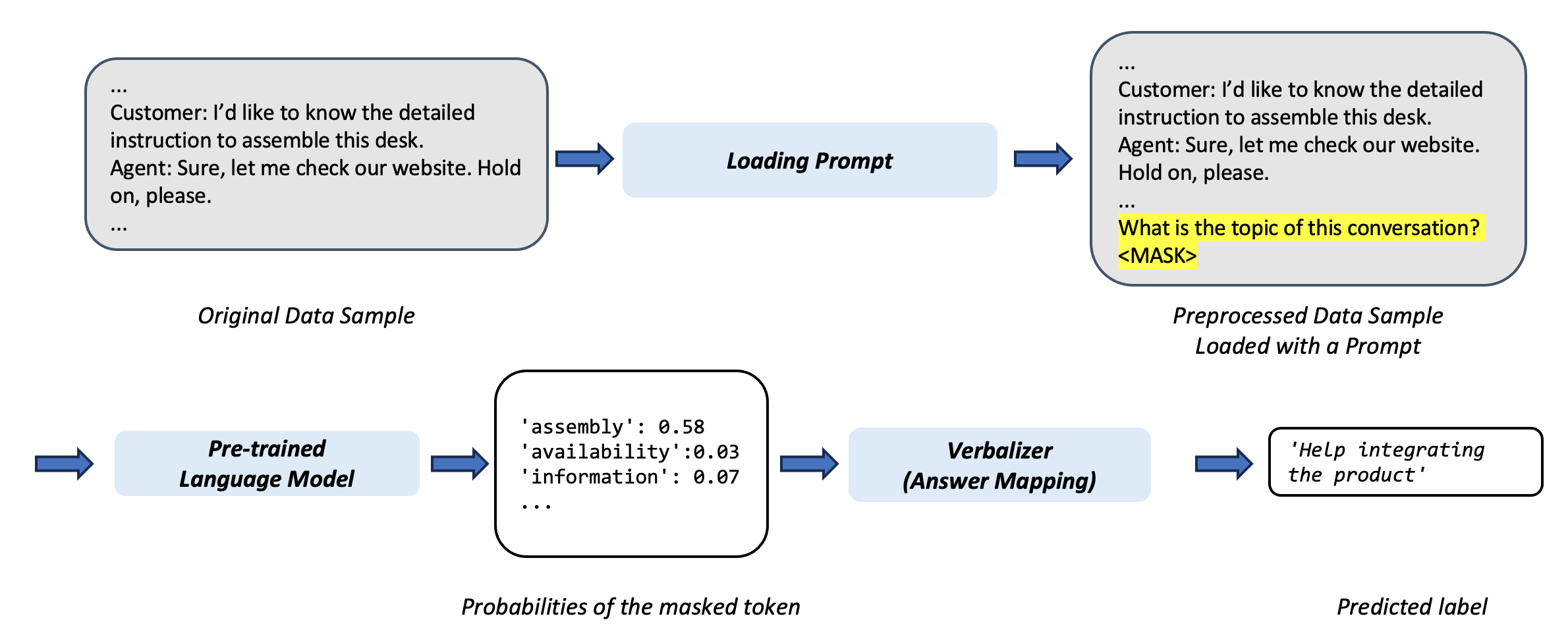}
    \caption{A demonstration of prompt-based learning pipeline}
    \label{pipeline}
\end{figure*}

The prompt-based learning pipeline, mathematically described by \citet{liu2023pre}, is a systematic process illustrated in Fig. \ref{pipeline}.

\subsubsection{Load Data with a Prompt}

The initial step involves augmenting the input text $x$ with a contextually relevant prompt to guide the language model's response generation. For instance, in a customer service setting with the input $x = $ ``... Customer: I’d like to know the detailed instruction to assemble this desk. Agent: Sure, let me check our website. Hold on, please. ...'', the prompt ``What is the topic of this conversation? <MASK>'' is appended, where `<MASK>' is the token to be predicted, thus transforming $x$ into $x' = $ ``Customer: I’d like to know the detailed instruction to assemble this desk. Agent: Sure, let me check our website. Hold on, please. What is the topic of this conversation? <MASK>''. We design various prompt templates in our prompt-based learning pipeline. For a detailed overview of the prompt templates we experimented with, see Appendix \ref{appendix:prompts-verbalizers}.

\subsubsection{Predict with a Language Model}

In this stage, the objective is to identify the token $\hat{z}$ that has the highest likelihood from the language model to fill in <MASK>. This process requires the definition of a set of allowable tokens for $z$, represented as $Z$. For instance, in this scenario, $Z$ might consist of potential topics such as \{``assembly'', ``availability'', ``information''\}, based on the context of the conversation. The optimal answer to fill in <MASK> token is then determined by computing the probability of all permissible tokens, seeking to maximize the predicted probability provided by the language model.

\subsubsection{Answer Mapping with a Verbalizer}

The final step entails converting the highest-scoring answer $\hat{z}$ to the corresponding output $\hat{y}$. In this example, the process involves mapping the most probable topic 'assembly' to a more actionable or interpretable label, such as ``Help integrating the product". This mapping is done using a verbalizer function that systematically correlates the answer space, composed of potential topic terms, to an output space of different labels that describe the customer's intent. We design various verbalizers in our prompt-based learning pipeline. For a detailed overview of the verbalizers we experimented with, see Appendix \ref{appendix:prompts-verbalizers}.

\subsection{Investigation in Prompt-based Learning Pipeline, with Prompted Training}

The performance of the prompt-based learning pipeline can be significantly influenced by the choice of language model, and the diversity of prompts and verbalizers used. In this context, we  do prompted training, which is to fine-tune language models within the pipeline using data that has been specifically formatted with prompts. This method leverages the inherent instruction-following capabilities of small language models (SLMs), which can deliver impressive results even when only a limited amount of data is available.

To further enhance the performance of this training approach, in this section, we explore advanced strategies of active data sampling, and the ensemble use of prompts and verbalizers.

\subsubsection{Active Data Sampling}

Instead of random data sampling, our goal is to probe the merits of selecting the most representative samples for each label. To achieve this, BERT \citep{devlin2018bert} is employed to embed text data from the training and development set into 512-dimensional vectors. Subsequently, we select samples that lie near the centroid of each class's sample cluster in the vector space. These centric samples which are likely to exemplify their respective classes, are speculated to raise the performance.

\subsubsection{Prompt and Verbalizer Ensemble}

Equipped with four templates and an equal number of verbalizers, we drew inspiration from PET \citep{schick-schutze-2021-exploiting} to adopt the concept of prompt ensemble, verbalizer ensemble, and prompt and verbalizer ensemble. To elaborate on our approach, for example, in prompt ensemble, we first assemble all short templates (templates No. 1-4) with a shared verbalizer, resulting in the creation of four distinct models using different templates but the same verbalizer. We then combine these four models by summing up the softmaxed logits and base our decision on the aggregated predicted probabilities. We follow a similar principle when assemble all the verbalizers together. Moreover, for prompt and verbalizer ensemble, we assemble a total of 16 models by combining the four templates and four verbalizers, allowing us to make a collective decision based on the combined outputs of these models.

\subsubsection{Baseline Settings}

We establish a baseline for convenience in the experimental settings and proceed to conduct experiments by modifying the components within the baseline pipeline. The baseline settings is shown in Table \ref{Baseline-tuning}.

\begin{table*}[htbp]
\centering
\begin{tabular}{ll}
\hline
\textbf{Component}                & \textbf{Choice}       \\ \hline
Language model & T5-base\tablefootnote{https://huggingface.co/t5-base} (Seq2seq LM, 220M Parameters)     \\
Data sampling strategy           & 5\% data with stratified sampling, randomly per label                                  \\
Choice of template       & Template 1   \\
Choice of verbalizer     & Verbalizer 1 \\ \hline
\end{tabular}
\caption{Baseline of the investigation in prompt-based learning
pipeline, with prompted training}
\label{Baseline-tuning}
\end{table*}

\subsection{Investigation in Prompt-based Learning Pipeline, without Prompted Training (Zero-shot)}

While prompted training within the prompt-based learning pipeline significantly enhances the performance of language models, it is also critical to explore the efficacy of these models in a zero-shot context. This investigation focuses on the performance of small language models (SLMs) when no fine-tuning is applied, examining their intrinsic instruction-following ability  based solely on their pre-trained configurations. The diversity of the language model choice and the specificity of prompts play pivotal roles in this scenario.

\subsubsection{Choice of Language Model}

In the exploration in zero-shot settings, where no training data is involved, we experimented with different model architectures (masked language models, language models, and seq2seq language models) to evaluate their impact on the intent recognition performance, as is shown in the table \ref{LM-experiment1}.

\begin{table}[htbp]
\centering
\begin{tabular}{lll}
\hline
\textbf{Language Model}    & \textbf{Architecture}      &  \textbf{Param. \#} \\ \hline

Roberta-large\tablefootnote{https://huggingface.co/roberta-large}      & Encoder-only         & 354M             \\

GPT-2-large\tablefootnote{https://huggingface.co/gpt2-large}        & Decoder-only & 774M             \\
GPT3.5-turbo\tablefootnote{https://openai.com/blog/chatgpt, with API version '2023-03-15-preview'}        & Decoder-only & $\sim$ 154B             \\

T5-large\tablefootnote{https://huggingface.co/t5-large}           & Encoder-decoder        & 770M             \\

FLAN-T5-large\tablefootnote{https://huggingface.co/google/flan-t5-large}      & Encoder-decoder        & 770M             \\ \hline
\end{tabular}
\caption{Different language models selected in zero-shot settings}
\label{LM-experiment1}
\end{table}

\subsubsection{Choice of Prompt: Detailed Prompt Design}

In this part, we will explore the effectiveness of 
a detailed prompt, which is designed to explicitly introduce the task and provide descriptions of all different labels to help the model better understand the task, as is shown in template 5 in Table \ref{prompt-template} in Appendix \ref{appendix:prompts-verbalizers}.

\subsubsection{Baseline Settings}

We establish a baseline for convenience in the experimental settings and proceed to conduct experiments by modifying the components within the baseline pipeline. We select verbalizer 1 In the baseline settings\footnote{Except experiments with GPT-3.5-turbo, as one of base models of ChatGPT. Since the model is packed in the OpenAI ChatGPT API and mapping the label with the verbalizer is not feasible, we add the prompt, “Return the index of the label, please.” and parse the response afterwards, instead of using a verbalizer.} .

\subsection{Evaluation Metrics}

We evaluate the effectiveness of the prompt-based learning pipeline with classic text classification metrics, which are accuracy and macro F1-score. Accuracy provides a direct measure of model performance by comparing the number of correct predictions with the total number of samples. On the other hand, macro F1-score takes into account the performance of the model for each class individually, making it another reliable metric especially for imbalanced datasets. Detailed definitions and formulas for these metrics can be found in the Appendix \ref{appendix:metrics-details}.

\subsection{OpenPrompt: An Open-Source Framework for prompt-based learning}

For our research, we employed OpenPrompt\footnote{https://github.com/thunlp/OpenPrompt}\citep{ding2021openprompt}, a dedicated open-source platform tailored for prompt-based learning studies. OpenPrompt offers an adaptable environment for the configuration and execution of diverse prompt-based learning components, ranging from prompt templates and verbalizers to pre-trained language models. Significantly, it ensures smooth integration with the Hugging Face Model Hub, streamlining the use of a variety of pre-trained language models.
Throughout this paper, OpenPrompt served as our primary tool for organizing and conducting prompt-based learning experiments.

\section{Experimental Results and Analysis}

\subsection{Investigation in Prompt-based Learning Pipeline, with Prompted Training}

\subsubsection{Prompted Training Compared with Traditional Fine-tuning}

We conduct experiments on the performance of prompted training compared to traditional fine-tuning approaches where no prompts are involved. The results are presented in Table \ref{General-comparison}.

\begin{table}[htbp]
\centering
\begin{tabular}{lll}
\hline
\textbf{Training Strategy}                  & \textbf{Accuracy} & \textbf{Macro F1} \\ \hline
100\% data, no prompt  & 80.63    & 72.84          \\
5\% data, no prompt   & 32.64    & 16.27          \\
5\% data, prompted    & \textbf{59.02}    & \textbf{45.49}       \\ \hline
\end{tabular}
\caption{Comparison of experimental results between prompted training and traditional fine-tuning}
\label{General-comparison}
\end{table}

The observations from Table \ref{General-comparison} demonstrate that the prompted training shows satisfactory performance compared to traditional fine-tuning approaches. This equivalence is particularly pronounced when working with limited labeled data. For instance, using only 5\% of labeled data, traditional fine-tuning witnesses a dramatic decline in accuracy, plummeting from 80.63\% to 32.64\%. In contrast, prompt-based fine-tuning retains higher accuracy and macro F1-score with the same data fraction.

Our extended experiments, detailed in Table \ref{different proportion}, indicate that increasing the proportion of labeled data sampled enhances the performance of the prompted training. Specifically, accuracy surges to 73.35\% and macro F1-score reaches 62.80\% as the sampled data grows from 1\% to 15\%.

\begin{table}[htbp]
\centering
\begin{tabular}{lll}
\hline
\textbf{Sampling Proportion} & \textbf{Accuracy} & \textbf{Macro F1} \\ \hline
1\%                        & 37.29    & 29.99          \\
3\%                        & 53.07    & 40.41          \\
5\%  (Baseline)                      & 59.02    & 45.49          \\
10\%                       & 67.84    & 55.80          \\
15\%                       & 73.35    & 62.80          \\ \hline
\end{tabular}
\caption{Experimental results of prompted training with different data sampling proportion}
\label{different proportion}
\end{table}

These results highlight the potential of the prompt-based learning pipeline with SLMs for sparsely labeled data scenarios. The probable reason might be the susceptibility of these models to overfitting when fine-tuned on smaller datasets.

Traditional fine-tuning methods often require substantial amounts of data to adapt pretrained models to specific downstream tasks. According to \citet{liu2023pre}, this is because there's a significant "gap" between training objectives, or in other words, the settings of loss functions, of pre-training and the downstream tasks these models are fine-tuned for. On the other hand, prompt-based learning seeks to bridge this gap. Instead of heavily adapting the model's parameters, prompt-based learning utilizes the vast knowledge already embedded in pretrained models, transforming downstream tasks to fit the pre-training objectives. By specific using prompts, this method translates tasks into a natural language format that the model is familiar with, effectively mining the pretrained model's potential. 

Furthermore, traditional fine-tuning often introduces new parameters for downstream tasks (e.g. introducing a classification head), whereas prompt-based learning maintains the same objectives as the pre-training phase and doesn't require additional parameters. This approach reduces the risk of overfitting, especially in limited data scenarios, making it a more data-efficient strategy.

Based on the promising results shown above, by modifying various components within this pipeline, we expect that optimal utilization of limited labeled data is achievable, leading to augmented performance.

\subsubsection{Active Data Sampling}

We conducted the experiments by varying the data sampling strategy from the baseline, to do random sampling or active sampling from the training and development set, and the experimental results are presented below in Table \ref{different-few-shot-active-sampling}.

\begin{table*}[htbp]
\centering
\begin{tabular}{lll}
\hline
\textbf{Data Sampling Strategy}            & \textbf{Accuracy} & \textbf{Macro F1-score} \\ \hline
3\% stratified sampling, randomly per label & 53.07    & 40.41          \\
3\% stratified sampling, actively per label & \textbf{60.03}    & \textbf{49.75}          \\
5\% stratified sampling, randomly per label (Baseline) & 59.02    & 45.49          \\
5\% stratified sampling, actively per label & \textbf{62.01}    & \textbf{52.77}          \\ \hline
\end{tabular}
\caption{Experimental results of prompted training with active or random data sampling strategies}
\label{different-few-shot-active-sampling}
\end{table*}

From Table \ref{different-few-shot-active-sampling}, it's evident that active sampling notably boosts accuracy and macro F1-score outperforming random sampling (7\% performance gain with 3\% data sampled, and 3\% performance gain with 5\% data sampled). This indicates that, a careful selection of dataset samples significantly uplifts performance over random data sampling.

Specifically, active learning, by design, prioritizes the most "informative" examples in the training set.  The idea is to maximize the information gain from each labeled instance, thus more efficiently updating the model's parameters. This strategy can be particularly impactful in scenarios with limited data to be trained.

\subsubsection{Prompt and Verbalizer Ensemble}

We conducted the experiments by varying the prompts and verbalizers from the baseline, and the experimental results are presented below in Table \ref{different-template-verbalizer}.
\begin{table*}[htbp]
\centering
\resizebox{\textwidth}{!}{%
\begin{tabular}{llllll}
\hline
\diagbox{\textbf{Template No.}}{\textbf{Acc. / Macro F1-score}} {\textbf{Verbalizer No.}} & \textbf{1}             & \textbf{2}             & \textbf{3}             & \textbf{4}             & \textbf{.*}             \\ \hline
1 & 59.02 / 45.49 & 60.56 / 53.66 & 61.95 / 50.89 & 60.67 / 48.20 & \textbf{64.74 / 54.63} \\
2 & 62.49 / 48.42 & 60.88 / 53.42 & 65.91 / 52.69 & 62.49 / 55.14 & \textbf{68.96 / 58.40} \\
3 & 60.62 / 45.58 & 62.38 / 55.89 & 64.04 / 49.19 & 60.78 / 52.51 & \textbf{67.04 / 55.23} \\
4 & 62.17 / 47.76 & 60.19 / 46.01 & 65.60 / 50.11 & 59.22 / 43.76 & \textbf{67.09 / 52.80} \\
.* & \textbf{64.63 / 49.84} & \textbf{63.77 / 55.98} & \textbf{67.79 / 53.60} & \textbf{63.02 / 52.73} & \textbf{68.70 / 56.76} \\ \hline
\end{tabular}%
}
\caption{Experimental results of prompted training with different templates and verbalizers (“.*” means all the templates / verbalizers are assembled), baseline with template 1 and verbalizer 1}
\label{different-template-verbalizer}
\end{table*}

From the experimental results shown in Table \ref{different-template-verbalizer}, we can see that, different templates or verbalizers may violate the effectiveness of the model and it shows a lack of robustness of prompt-based learning process. Some prompts may align better with certain types of input data than others, resulting in performance discrepancies. This variance underscores the importance of the aggregation with various prompts and verbalizers – a kind of "ensemble learning" that can enhance the robustness of the pipeline.

Accordingly, we can observe from the table \ref{different-template-verbalizer} that by aggregating templates and/or  verbalizers, the performance of the pipeline improves greatly and surpasses that of any individual model alone. This highlights the effectiveness of combining multiple templates and verbalizers to enhance the overall performance and achieve better intent recognition results in comparison to solely using one single template combined with one verbalizer.

\subsection{Investigation in Prompt-based Learning Pipeline, without Prompted Training (Zero-shot)}

We conducted the experiments by varying lanaguage models with a different prompt templates (template 1 and 5), and the experimental results are presented below in Table \ref{tuning-free-prompting-results}. Reviewing Table \ref{tuning-free-prompting-results}, short prompts, such as prompt 1, generally yield relatively low performance, consistently below 20\%, except the advanced GPT-3.5-turbo model. This limits their applicability in practical contexts.
\begin{table*}[htbp]
\centering

\begin{tabular}{lll}
\hline
\diagbox{\textbf{Name of Model (\# of Params)}}{\textbf{Acc. / Macro F1-score}}{\textbf{Template No.}}
                          & \textbf{1}             & \textbf{5}                         \\ \hline
RoBerta-large (354M)      & 4.07 / 1.81   & \textbf{11.66 / 7.91}    \\
GPT-2-large (774M)         & 3.96 / 1.66   & \textbf{7.01 / 3.74}     \\
T5-large (770M)           & 3.31 / 1.00   & \textbf{19.31 / 8.58}    \\
FLAN-T5-large (770M)      & 17.92 / 9.97  & \textbf{31.35 / 20.57}  \\ 
GPT-3.5-turbo ($\sim$154B)   & 31.57 / 19.57 & \textbf{55.16 / 43.38}  \\ \hline
\end{tabular}%

\caption{Experimental results without prompted training}
\label{tuning-free-prompting-results}
\end{table*}

This led us to explore detailed prompts as a potential enhancement. The outcomes, presented in Table \ref{tuning-free-prompting-results}, show significant performance elevations with a detailed prompt. For instance, the T5-large model's zero-shot accuracy leaped from 3.31\% to 19.31\%, and the FLAN-T5-large model's zero-shot accuracy rose from 17.92\% to 31.35\%. 
Besides, the GPT-3.5-turbo model's zero-shot accuracy surged past 50\%. This indicates that explicit, task-centric prompts can amplify performance in zero-shot scenarios. This demonstrates the importance of prompt design in zero-shot learning scenarios, when training data, or even demonstration data to put into prompts, is not available.

Moreover, even though GPT-3.5-turbo demonstrates state-of-the-art performance, Flan-T5-large's performance has the highest percentage increase, which nearly doubled in terms of accuracy, indicating strong instruction-following potential of SLMs in zero-shot settings.

\section{Conclusion and Future Work}

This paper has explored the efficacy of the prompt-based learning paradigm, specifically employing small language models (SLMs) for intent recognition in customer-agent conversations within the customer service domain. We have demonstrated that this paradigm significantly reduces the reliance on large labeled datasets typically required for traditional model training. Our findings reveal that even with minimal training data, SLMs can achieve competitive performance, with the utilization of well-designed prompts and strategic answer mapping techniques.

The integration of active sampling and ensemble learning strategies has further enhanced the performance of our prompt-based learning pipeline, offering robust solutions to the challenges of data sparsity and model efficiency in practical applications. Notably, our experiments in zero-shot settings have confirmed that SLMs, when equipped with detailed prompts, possess substantial instruction-following capabilities that allow them to handle complex customer intents without the need for extensive pre-training.

These results are particularly salient for advancing AI-driven customer service solutions, suggesting a shift towards more data-efficient, flexible, and cost-effective NLP models. The adoption of prompt-based learning with SLMs could improve the way businesses interact with their customers, providing more accurate service options based on comprehensive customer intent analysis.

Looking ahead, there are several promising avenues for further research. One potential area involves refining the methodologies (e.g. a more strategic verbalizer design, etc.) used in our prompt-based learning pipeline to enhance the adaptability and accuracy of SLMs across different domains. Besides, future work could explore the integration of multimodal data inputs to enrich the context and effectiveness of customer intent recognition systems.

\bibliography{custom}
\bibliographystyle{acl_natbib}

\appendix

\section{Detailed Evaluation Metrics}
\label{appendix:metrics-details}

\subsection{Accuracy}

Accuracy is defined as:
\begin{equation}
Accuracy = \frac{C}{T}
\end{equation}
Where \(C\) represents the number of correctly classified samples and \(T\) is the total number of samples in the dataset.

\subsection{Macro F1-score}

Macro F1-score is the average of the F1-scores for all the classes:
\begin{equation}
{Macro F1-score} = \frac{1}{N} \sum_{i=1}^{N} {F1-score}_i
\end{equation}
where \(N\) is the number of classes. The F1-score for each class, denoted as \({F1-score}_i\), is the harmonic mean of precision (\(P\)) and recall (\(R\)):
\begin{equation}
{F1-score}_i = \frac{2 \times P \times R}{P + R}
\end{equation}
Macro F1-score is at times preferred over accuracy because it considers each class equally, regardless of its size.

\section{Hyperparameter Settings}

We provide a comprehensive overview of the experimental environment and hyperparameters used with OpenPrompt in Table \ref{Experimental-Enviroment} and \ref{openprompt-Enviroment}.

\begin{table}[htbp]
\centering
\begin{tabular}{ll}
\hline
Environment & Detail                      \\ \hline
Platform                 & Google Cloud Platform \\
PyTorch version          & 1.13                        \\
Machine Type             & 16 vCPUs, 104 GB RAM        \\
GPUs                     & NVIDIA T4 * 2               \\ \hline
\end{tabular}
\caption{Experimental enviroment}
\label{Experimental-Enviroment}
\end{table}

\begin{table*}[htbp]
\centering
\begin{tabular}{ll}
\hline
OpenPrompt Hyperparameters                                                           & Settings \\ \hline
Random seed                                                                          & 144    \\
Batch size                                                                           & 5      \\
Training Epochs  & 20     \\
Training Strategy  &
  \begin{tabular}[c]{@{}l@{}}Select the checkpoint out of 20,\\ where accuracy on dev set is the highest.\end{tabular} \\
Optimizer        & AdamW  \\
Weight decay  &
  \begin{tabular}[c]{@{}l@{}}0.01 (except {[}“bias”, “LayerNorm.weight”{]});\\ 0 ({[}“bias”, “LayerNorm.weight”{]})\end{tabular} \\
Learning rate   & $3 \times 10^{-5}$   \\ \hline
\end{tabular}
\caption{OpenPrompt hyperparameters}
\label{openprompt-Enviroment}
\end{table*}

\section{Design of Prompt Template and Verbalizer}
\label{appendix:prompts-verbalizers}


The design of prompt template and verbalizer is a crucial step in the prompt-based learning pipeline. We designed four one-sentence prompt templates and one detailed prompt template that we will use in our experiments as is shown in the table \ref{prompt-template}. 

\begin{table*}[htbp]
\centering
\begin{tabular}{ll}
\hline
Template No. & Prompt Template                                                      \\ \hline
Template 1          & “\{conversation\} Classify this conversation : \{mask\}”             \\
Template 2          & “\{conversation\} What is the topic of this conversation ? \{mask\}” \\
Template 3          & “\{conversation\} What is the intent of the customer ? \{mask\}”     \\ 
Template 4          & “\{conversation\} We will be happy to help you with your  \{mask\}.”     \\ 
Template 5 &
  \begin{tabular}[c]{@{}l@{}}“\{conversation\} \\ Given this conversation,  we have 14 classes:\\ General: General information and issues customer has before buying;\\ Product / Service Information: Information about products and services;\\ ...\\ Please classify this conversation into one class out of these 14 classes: \{mask\}”\end{tabular} \\ \hline
\end{tabular}
\caption{Different prompt templates tested in the prompt-based learning pipeline}
\label{prompt-template}
\end{table*}

And besides, the verbalizer, as a crucial component of the prompt-based learning pipeline, maps the outputs of the language model to the desired labels or categories. We introduce four different verbalizers that we will use in our experiments, shown below in the Table \ref{different-verbalisers}.

\begin{table*}[htbp]
\centering
\resizebox{\textwidth}{!}{%
\begin{tabular}{lllll}
\hline
Label No.                     & Label words of Verbalizer 1               & Label Words of Verbalizer 2      & Label Words of Verbalizer 3   & Label Words of Verbalizer 4         \\ \hline
0                           & [“availability”]                      & [“availability”,“stock”,“order”]  & [“availability”,“purchase”]                      & [“availability”,“stock”,“order”,“purchase”]    \\
1                           & [“general”]                           & [“general”,“membership”]      & [“general”,“problem”]                           & [“membership”,“problem”,“account”]           \\
2                           & [“general”, “purchase”]              & [“general”,“help”]    & [“general”, “purchase”,“problem”]              & [“general”,“help”,“problem”]        \\
3                           & [“help”,“integrate”,“product”]        & [“assembly”,“product”]        & [“help”,“integrate”,“product”,“purchase”]        & [“assembly”,“product”,“purchase”]         \\
4                           & [“initiate”, “sales”]              & [“aftersales”]   & [“initiate”, “sales”,“problem”]              & [“aftersales”,“problem”]         \\
5                           & [“issue”, “handling”]              & [“issue”, “refund”]  & [“issue”, “handling”,“refund”]              & [“issue”, “refund”]          \\
6                           & [“order”, “creation”]             & [“order”, “availability”,“delivery”]   & [“order”, “creation”]              & [“order”, “availability”,“delivery”]         \\
7                           & [“order”, “fulfillment”, “issues”]             & [“order”, “product”, “refund”,“fulfillment”]  & [“order”, “fulfillment”, “issue”,“problem”]              & [“order”, “product”, “refund”,“problem”]          \\
8                           & [“order”, “processing”]             & [“order”, “address”,“delivery”]  & [“order”, “processing”]              & [“order”, “address”,“delivery”]          \\
9                           & [“other”] & [“other”] & [“other”] & [“other”] \\
10                          & [“planning”, “advice”]              & [“planning”, “advice”,“suggestion”]  & [“planning”, “advice”,“project”]              & [“planning”, “advice”,“suggestion”,“project”]          \\
11                          & [“prepare”, “exchange”,“return”]            & [“exchange”,“return”]   & [“prepare”, “exchange”,“return”]            & [“exchange”,“return”]         \\
12                          & [“product”, “service”, “information”] & [“stock”,“delivery”,“information”,“order”] & [“product”, “service”, “information”,“order”] & [“stock”,“delivery”,“information”,“order”] \\
13                          & [“service”, “fulfillment”]              & [“service”, “fulfillment”]           & [“service”, “fulfillment”,“order”]           & [“service”, “fulfillment”,“order”]        \\ \hline
\end{tabular}%
}
\caption{Different verbalizers tested in the prompt-based learning pipeline}
\label{different-verbalisers}
\end{table*}

The first verbalizer harnesses the phrasing of original intention labels set in the dataset. The second verbalizer extracts label words from the text via the unsupervised topic modeling tool BERTopic \footnote{https://maartengr.github.io/BERTopic} \citep{grootendorst2022bertopic} coupled with manual scrutiny. Further, we automatically generated the label words based on prompt template 4, following the methodology of LM-BFF \citep{gao-etal-2021-making}. These words then added the first two verbalizers, forming the third and fourth variants.

\end{document}